\documentclass[letterpaper]{article} 
\usepackage{aaai2026}  
\usepackage{times}  
\usepackage{helvet}  
\usepackage{courier}  
\usepackage[hyphens]{url}  
\usepackage{graphicx} 
\usepackage{amssymb}
\usepackage{natbib}  
\usepackage{caption} 
\usepackage{algorithm}
\usepackage{algorithmic}
\usepackage{amsmath}
\usepackage{multirow}

\usepackage{newfloat}
\usepackage{listings}
\DeclareCaptionStyle{ruled}{labelfont=normalfont,labelsep=colon,strut=off} 
\floatstyle{ruled}
\newfloat{listing}{tb}{lst}{}
\floatname{listing}{Listing}
\title{GIIM: Graph-based Learning of Inter- and Intra-view Dependencies for Multi-view Medical Image Diagnosis}
\author{
    Tran Bao Sam,
    Hung Vu,
    Dao Trung Kien,
    Tran Dat Dang,
    Van Ha Tang,
    Steven Truong
}
\affiliations{
    NVIDIA

    Ho Chi Minh City, Vietnam\\
    sbtran@nvidia.com
}

\begin{document}

\maketitle

\begin{abstract}
Computer-aided diagnosis (CADx) has become vital in medical imaging, but automated systems often struggle to replicate the nuanced process of clinical interpretation. Expert diagnosis requires a comprehensive analysis of how abnormalities relate to each other across various views and time points, but current multi-view CADx methods frequently overlook these complex dependencies. Specifically, they fail to model the crucial relationships within a single view and the dynamic changes lesions exhibit across different views. This limitation, combined with the common challenge of incomplete data, greatly reduces their predictive reliability. To address these gaps, we reframe the diagnostic task as one of relationship modeling and propose GIIM, a novel graph-based approach. Our framework is uniquely designed to simultaneously capture both critical intra-view dependencies between abnormalities and inter-view dynamics. Furthermore, it ensures diagnostic robustness by incorporating specific techniques to effectively handle missing data, a common clinical issue. We demonstrate the generality of this approach through extensive evaluations on diverse imaging modalities, including CT, MRI, and mammography. The results confirm that our GIIM model significantly enhances diagnostic accuracy and robustness over existing methods, establishing a more effective framework for future CADx systems.
\end{abstract}


\section{Introduction}
\label{sec:intro}

Medical image classification is a fundamental component of medical image analysis, crucial for supporting accurate diagnoses and effective treatment planning across a variety of imaging modalities like computed tomography (CT), magnetic resonance imaging (MRI), X-ray and mammography.
Initially, the field relied on traditional classification methods using predefined and handcrafted feature extraction. While useful, these techniques were often limited in their flexibility and adaptability. An advanced example of this feature-based approach is radiomics \cite{lambin2012radiomics}, which extracts detailed statistical and gray-level information from regions of interest (ROIs) and has shown promise when paired with machine learning predictors \cite{wu2019radiomics}, \cite{liu2021can}, \cite{severn2022pipeline}, \cite{ha2023radiomics}.
A significant transformation came with the arrival of deep learning techniques, especially convolutional neural networks (CNNs). This paradigm shift introduced automatic feature extraction, leading to substantial performance improvements in a wide range of diagnostic tasks. CNNs have demonstrated exceptional efficacy in classifying complex patterns, such as identifying interstitial lung disease in chest CT scans \cite{anthimopoulos2016lung}, \cite{peter2023} or distinguishing between different types of brain tumors \cite{abiwinanda2019brain}, \cite{ayadi2021deep}, \cite{ullah2024} and early detecting abnormalities in mammography \cite{luo2024}, \cite{naga2024}.

\begin{figure*}[!ht]
    \centering
    \includegraphics[width=0.94\textwidth]{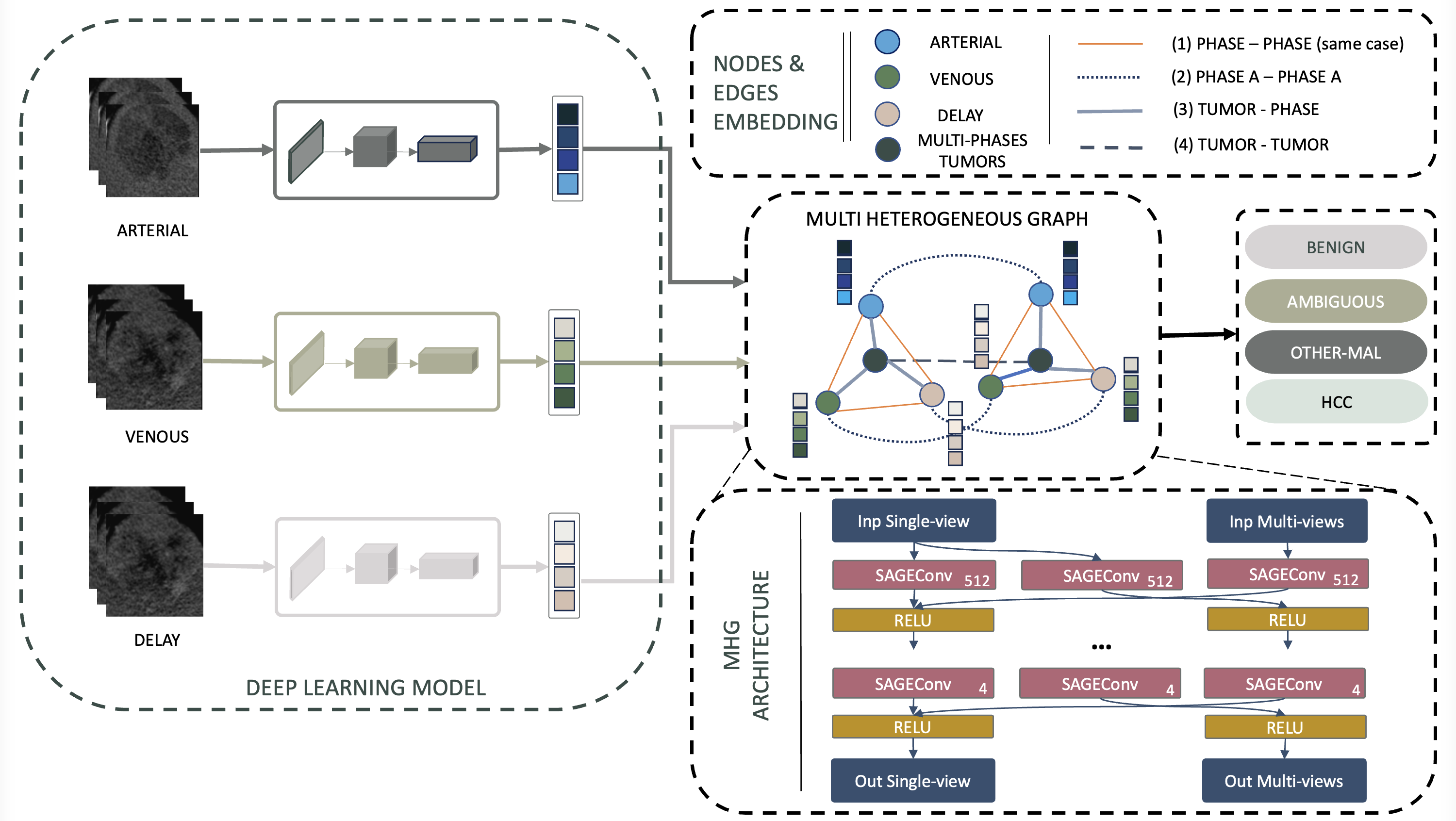}
    \caption{Illustration of our approach for focal liver tumor classification. Two main components consist of (i) the extraction of single-view features and (ii) GIIM.}
    \label{fig:liver_arch}
\end{figure*}

Alongside CNNs, transformer-based architectures like attention modules and Vision Transformers (ViTs) have shown strong potential in medical imaging \cite{wang2022phase}, \cite{fran2025}. For example, \cite{wang2022phase} introduced a model with intra-phase and inter-phase attention, enabling focus on key spatial locations within a view and tracking changes across phases to capture temporal dynamics. Recently, Graph Neural Networks (GNNs) \cite{scarselli2008graph} and their heterogeneous variants \cite{wang2019heterogeneous} have become powerful tools for modeling complex dependencies in perceptual data. GNNs excel at representing multifaceted relationships and diverse entity types, leading to improved performance. Several recent works have applied GNNs to medical diagnosis, such as SSL-MNGCN and MMGCN \cite{ibrahim2022multi}, which classify mammography images by modeling texture and spatial features as graphs. GNNs have also been used to contextualize connections between different body parts in lung CT scans, enhancing the understanding of complex anatomical relationships \cite{sun2022context}.

Within the realm of medical image analysis, multi-view images have become indispensable, offering complementary perspectives that enrich the information available to radiologists. The utilization of multiphase CT volumes \cite{sam2024}, \cite{hung2025} enables the monitoring of changes in organ intensity over time, providing critical insights into patient health. Similarly, multi-view mammography \cite{nguyen2022novel} improves tumor detection by capturing images from different angles, reducing the chances of missing lesions hidden by overlying tissues. Some methods leverage these multiple views by first aligning the images and then combining them into a single, fused image \cite{sharif2021decision}, \cite{ouhmich2019liver}. These techniques often use separate CNN encoders for each view to extract features, which are then fused through strategies like addition, multiplication, or averaging for classification tasks. Additionally, pretrained models, such as multiphase convolutional dense networks, have been developed to analyze four-phase mammograms, leading to improved lesion detection from multiphase data \cite{cao2020multiphase}, \cite{wang2022phase}.

A comprehensive clinical diagnosis requires a holistic assessment of how abnormalities relate to each other. Factors such as tumor size, location, and spatial relationships between multiple tumors are critical in cancer diagnosis \cite{lee2025}, \cite{wu2018}, \cite{zhang2020}. Despite advances in CADx, many existing methods, including those based on ML, CNNs, transformers, and GNNs, typically analyze lesions independently, overlooking crucial diagnostic clues from interdependencies among abnormalities. They fail to model relationships within a single view (intra-view) and changes across different views or time points (inter-view dependencies). This limitation is particularly problematic for diagnosing small or ambiguous tumors. To address these gaps, we introduce our proposed method, \textbf{GIIM}, as it is a \textbf{G}raph-based approach that models both \textbf{I}nter- and \textbf{I}ntra-view dependencies for \textbf{M}ulti-view medical image classification. Our novel architecture, based on Multi-Heterogeneous Graphs (MHGs), captures complex relationships and correlations between all lesions of a patient. This graph-based paradigm offers advantages over methods like CNNs and Transformers, which require fixed-size inputs. GNNs are inherently flexible, modeling a variable number of lesions and their intricate connections, making GIIM framework suitable for clinical diagnosis. GIIM also mitigates issues associated with incomplete multi-view data, which can negatively affect diagnostic performance. Multi-view imaging provides comprehensive information but often encounters missing data issues, hindering image-based analysis and raising uncertainties in data classification \cite{Patel2020c}, \cite{zhang2022mmformer}, \cite{xie2023exploring}. Addressing these limitations significantly enhances medical image classification performance and potentially aids in clinical practice. 

The main contributions can be highlighted as follows.

\begin{enumerate}
    \item \textbf{A Novel GIIM Architecture:} We propose GIIM architecture based on MHGs that is uniquely designed to integrate both intra-view and inter-view structural dependencies. This breakthrough enhances diagnostic accuracy across diverse imaging modalities, including CT, MRI, and mammography.

    \item \textbf{Robustness to Incomplete Data:} Our approach effectively addresses the common challenge of incomplete multi-view medical images. We introduce four techniques for missing data representation to ensure robust performance in missing-view scenarios.

    \item \textbf{Extensive Experimental Validation:} The proposed model is evaluated on multiple real medical datasets under both complete and incomplete data settings. Through extensive experiments, our GIIM significantly improves diagnostic outcomes and outperforms existing methods in accuracy and area under the receiver operating characteristic curve (AUC).

\end{enumerate} 

\begin{algorithm}[tb]
\caption{GIIM Training Pipeline}
\label{alg:training}
\textbf{Input}: Dataset $D$ of all tumor images, grouped by patient. \textbf{Output}: $V$ trained feature extractors $\{f_v\}_{v=1}^V$ and a trained MHG model $\mathcal{M}_{MHG}$.

\begin{algorithmic}[1] 
\STATE \textbf{Stage 1: Train Single-View Feature Extractors}
\FOR{$v = 1 \to V$}
    \STATE Let $D_v$ be the dataset of all tumor images for view $v$.
    \STATE Initialize single-view model $N_v$ (e.g., ConvNeXt).
    \FOR{each training epoch}
        \FOR{each tumor sample $(x_{tumor}, y_{tumor})$ in $D_v$}
            \STATE Predict $\hat{y} \gets N_v(x_{tumor})$.
            \STATE Calculate loss $\mathcal{L}(\hat{y}, y_{tumor})$.
            \STATE Update parameters of $N_v$.
        \ENDFOR
    \ENDFOR
    \STATE Freeze the trained backbone of $N_v$ and denote it as feature extractor $f_v$.
\ENDFOR

\STATE
\STATE \textbf{Stage 2: Train the MHGs}
\STATE Group dataset by patient: $D_{\text{patient}} = \{(P_i, Y_i)\}_{i=1}^{\text{NumPatients}}$, where $Y_i$ is the set of all tumor-level labels for patient $i$.
\STATE Initialize the MHG model $\mathcal{M}_{MHG}$.
\FOR{each training epoch}
    \FOR{each patient case $(P_i, Y_i)$ in $D_{\text{patient}}$}
        \STATE Let $\text{PatientFeatures} \leftarrow []$.
        \FOR{each lesion $l$ in patient $P_i$}
            \STATE Extract features for all its views: $N_{single} = \{f_v(x_{l,v}) \mid v=1 \dots V\}$.
            \STATE Append $N_{single}$ to $\text{PatientFeatures}$.
        \ENDFOR
        \STATE \COMMENT{Construct a HG for patient $i$ using Eq~\ref{equa:0},\ref{equa:1},\ref{equa:2},\ref{equa:3}. }
        \STATE $\mathbb{G}_i \gets \text{ConstructMHG}(\text{PatientFeatures})$.
        \STATE Let $Z_i \gets \mathcal{M}_{MHG}(\mathbb{G}_i)$. \COMMENT{$Z_i$ contains output logits for each tumor node.} using Eq.\ref{equa:5},\ref{equa:6},\ref{equa:7}.
        \STATE Calculate loss $\mathcal{L}(Z_i, Y_i)$ over all tumor nodes.
        \STATE Update parameters of model $\mathcal{M}_{MHG}$.
    \ENDFOR
\ENDFOR
\STATE \textbf{return} $\{f_v\}_{v=1}^V$ and $\mathcal{M}_{MHG}$
\end{algorithmic}
\end{algorithm}

\section{Proposed Method}
\label{sec:method}
This section presents the proposed method that comprises two main parts: (i) the extraction of single-view features and (ii) GIIM as a comprehensive multi-view approach. The visual representation of our approach is illustrated in Figure ~\ref{fig:liver_arch}. The pseudo code of the proposed method is also presented in Algorithm 1. 

\subsection {Single View Feature Extraction}

The single-view network architecture is designed to automatically extract features from images, which are then mapped to predictions via two fully connected layers. For this task, we adopt the ConvNeXt \cite{convnext} architecture as our feature extractor. ConvNeXt \cite{convnext} is particularly well-suited for medical image analysis because its modern convolutional design effectively captures both the large-scale context and fine-grained details crucial for accurate diagnosis. Its use of large-kernel spatial convolutions \( (7 \times 7) \) provides a larger receptive field to better analyze the morphology of sizable tumors and understand the spatial relationship between a lesion and surrounding tissue. Additionally, its efficient depthwise separable convolutions are adept at learning detailed spatial patterns, which helps maintain lesion boundaries and sensitivity to changes in tumor intensity. Our methodology involves individually training distinct models on their corresponding single-view datasets. For a dataset with $V$ views, this process generates $V$ separate models, each tailored to a specific view to produce meaningful features. These extracted features are then harnessed as inputs for the subsequent multi-view model, enabling a comprehensive analysis of the data's underlying correlations.

\subsection{GIIM: Graph-based Model of Inter and Intra-view Dependencies for Multi-view}

This section details how we represent multi-view medical data as a heterogeneous graph and the message-passing mechanism used for learning. We denote the total number of samples in the dataset as $S$. Each sample, which contains $V$ different views, is represented as a heterogeneous graph $\mathbb{G}_s$ from the collection $\{\mathbb{G}_s\}_{s=1}^S$. Each graph $\mathbb{G}_s$ is composed of a set of vertices (nodes) $N$ and a collection of edges $E$, and is associated with a class label $y_s$.

\subsubsection{Node Representation}

Two primary types of nodes exist within each graph:
\begin{itemize}
    \item \textbf{Single-view node ($N_{single}$):} A set of nodes $\{N_{single_v}\}_{v=1}^V$, where each node represents one of the $V$ distinct views.
    \item \textbf{Multi-view node ($M_{multi}$):} A single node created by concatenating the features of all single-view nodes: $M_{multi} = \Vert_{v=1}^V (N_{single_v})$.
\end{itemize}

\subsubsection{Edge Representation}

The set of edges $E$ models the relationships between nodes. Let $N_{v,j}$ be the node for single-view $v$ of lesion $L_j$, and $M_j$ be the multi-view summary node for lesion $L_j$. The total edge set $E$ is the union of four distinct subsets:

\begin{itemize}
    \item \textbf{Intra-tumor, Inter-view ($E_{intra}$):} Connects different views of the same lesion to capture temporal changes.
        \begin{equation}\label{equa:0}
            E_{intra} = \{(N_{v,j}, N_{v',j}) \mid j \in L, v \neq v'\}
        \end{equation}

    \item \textbf{Single-to-Multi-view ($E_{s-m}$):} Connects a single-view node to its corresponding multi-view summary node.
        \begin{equation}\label{equa:1}
            E_{s-m} = \{(N_{v,j}, M_j) \mid j \in L, v \in V\}
        \end{equation}

    \item \textbf{Inter-tumor, Single-view ($E_{inter-s}$):} Connects different lesions observed in the same view.
        \begin{equation}\label{equa:2}
            E_{inter-s} = \{(N_{v,j}, N_{v,j'}) \mid v \in V, j \neq j'\}
        \end{equation}

    \item \textbf{Inter-tumor, Multi-view ($E_{inter-m}$):} Models the high-level contextual relationship between all lesions in the case, which is useful for identifying small lesions by their proximity to larger ones.
        \begin{equation}\label{equa:3}
            E_{inter-m} = \{(M_j, M_{j'}) \mid j \neq j'\}
        \end{equation}
\end{itemize}

\subsubsection{Heterogeneous Message Passing}

To learn from this structure, we employ a heterogeneous message-passing scheme where the feature update for any node $n$ at layer $k$ separately processes its neighbors based on their type.

\paragraph{1. Neighbor Aggregation}
We compute two distinct aggregated feature vectors for node $n$: one from its single-view neighbors and one from its multi-view neighbors.

\begin{itemize}
    \item Aggregation from Single-view neighbors (${N}_{single}(n)$):
        \begin{equation}\label{equa:5}
            h_{{N}_{single}(n)}^k \leftarrow \frac{1}{|{N}_{single}(n)|} \sum_{u \in {N}_{single}(n)} \mathbf{W}_{single}^k h_u^{k-1}
        \end{equation}
    \item Aggregation from Multi-view neighbors (${M}_{multi}(n)$):
        \begin{equation}\label{equa:6}
            h_{{M}_{multi}(n)}^k \leftarrow \frac{1}{|{M}_{multi}(n)|} \sum_{u \in {M}_{multi}(n)} \mathbf{W}_{multi}^k h_u^{k-1}
        \end{equation}
\end{itemize}
Here, $\mathbf{W}_{single}^k$ and $\mathbf{W}_{multi}^k$ are distinct, trainable weight matrices for layer $k$ that learn different transformations for each neighbor type.

\paragraph{2. Final Update}
The final feature vector $h_n^k$ is computed by concatenating the node's own previous state, $h_n^{k-1}$, with the two aggregated neighborhood vectors. This combined vector is then passed through a final dense layer with a non-linear activation function $\sigma$.
\begin{equation}\label{equa:7}
    h_n^k \leftarrow \sigma \left( \mathbf{W}^k \cdot \text{CONCAT}(h_n^{k-1}, h_{{N}_{single}(n)}^k, h_{{M}_{multi}(n)}^k) \right)
\end{equation}
Where $\mathbf{W}^k$ is the main weight matrix for the update step in layer $k$.

\subsubsection{MHG Architecture} In terms of architecture design, the proposed MHG-based model contains a sequence of five SAGEConv layers with varying units of 512, 256, 128, 64, and 32. The final SAGEConv layer has a number of units equivalent to the output classes, used to compute the prediction probability, thus supplanting the need for a fully connected layer. A detailed illustration of the model architecture is provided in Figure ~\ref{fig:liver_arch}.  In the following, we present the proposed GIIM for typical multi-phase and multi-view medical imaging diagnosis problems. 

\subsection{GIIM in Medical Context}
\subsubsection{Liver Tumor Classification on Multi-phase CT}

Accurate classification and characterization of focal liver lesions (FLLs) are vital in guiding effective patient treatment strategies. Liver examination is commonly conducted using various medical imaging techniques, including CT, MRI, ultrasound, and X-ray. Among these, multi-phase contrast-enhanced CT emerges as an effective approach for detecting FLLs, particularly those related to hepatocellular carcinoma (HCC). Such FLLs can be categorized by processing the multi-phase CT scans comprising arterial, venous, and delay phases. 

\begin{figure}[!htpb]
\centering 
    \includegraphics[width=0.44\textwidth]{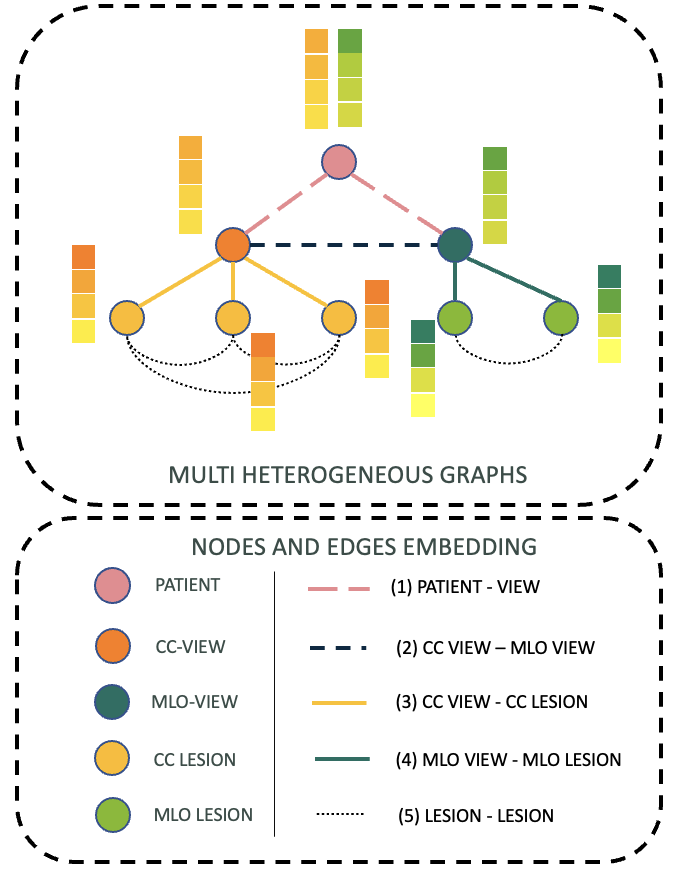}
    \caption{Our proposed GIIM for BI-RADS classification.}
    \label{fig:mammoGMHs}
\end{figure}

The proposed GIIM model is capable of harnessing the distinctive features presented across various phases for classifying the FLLs.
Let $S$ denote the number of patients present in the dataset. Each patient is associated with $L$ lesions, 
and every lesion is observed across three distinct phases denoted by $V = \{Arterial, Venous, Delay\}$. Our GIIM, denoted as $\mathbb{G}_{MHGs} = \{ \mathbb{G}(N, E)\}_{s=1}^{S}$,  comprises node embeddings $ \{N = \{{N}_{single}, {M}_{multi}\}\}_{l=1}^{L} $. 

For FLL classification, the edge set $E$ models four key relationships: $E_{intra}$ capture temporal changes within a single lesion, while $E_{s-m}$ integrate these phase-specific details into a holistic summary. To model context between different lesions, $E_{inter-s}$ correlate them at specific time points, and $E_{inter-m}$ connect their overall summary nodes. This inter-lesion modeling is particularly beneficial as it provides contextual clues, acknowledging that certain tumor types often appear together.

\subsubsection{BIRADS Classification on Mammograms}
Breast cancer has now taken the lead as the most prevalent cancer among women, as confirmed by statistics from the International Agency for Research on Cancer (IARC) in December 2020 \cite{iarc}. Mammography, a widely adopted X-ray imaging technique for breasts, plays a crucial role in CADx systems aimed at supporting radiologists in evaluating breast cancer risk.  For a patient, there are two distinct breast sides: left and right, each with two views - CC (top-down) and MLO (side). Within these views, $L$ lesions are present. CADx in mammograms faces challenges when a lesion in the CC view might not be visible in the MLO view. Mapping lesion A from the CC view to lesion B in the MLO view becomes complicated due to the lack of necessary information and differing observation angles. Therefore, GIIM designed for BI-RADS classification slightly differ from GIIM for FLLs classification. The single-view node embedding $N_{v} = \frac{1}{L} \sum_{l=1}^{L} f_v(l_v)$ represents the mean of all lesion features within that specific view, while the multi-view node embedding $M_{m} = \Vert^V_{v = 1} M_v$ signifies the concatenation of different single-view features. The detailed architecture of GIIM for BI-RADS classification is illustrated in Figure \ref{fig:mammoGMHs}.

\begin{table*}[h]
\centering

\begin{tabular}{cccccccc}
\hline
\multirow{2}{*}{Group} & \multicolumn{2}{c}{Data} & \multicolumn{5}{c}{Number of Tumors} \\ \cline{4-8}
& \# Patients & \# Samples & Benign & Ambiguous & Malignant & HCC & Total \\ \hline
Train & 625 & 659 & 561 & 195 & 147 & 491 & 1394 \\ \hline
Test & 252 & 261 & 176 & 43 & 56 & 165 & 440 \\ \hline
Total & 877 & 920 & 737 & 238 & 203 & 656 & 1834 \\ \hline
\end{tabular}
\caption{The statistics of the liver tumor dataset.}
\label{tab:liverdataset}
\end{table*}

\subsubsection{Breast Lesions Classification on MRI}

In this scenario, $V = \{\textit{Pre-contrast}, \textit{Post-contrast}, \textit{Subtraction}\}$ was defined as a set representing the three distinct MRI sequences acquired for each case study. A function $f_v(\cdot)$ for $v \in V$ represents the feature extractor for each sequence. This function extracts relevant information from the input data $l_v$ associated with sequence $v$. The proposed GIIM is denoted as $\mathbb{G}_{MHGs} = \{ \mathbb{G}(N, E) \}$. Here, $\mathbb{G}(N, E)$ represents a graph for each case study, where $N$ represents the set of nodes and $E$ represents the edges connecting them. The GIIM model utilizes two types of node embeddings for each patient: (1) Single-Sequence Features $N_{single} = f_v(l_v)$ represents the features extracted from sequence $v$ using the corresponding feature extractor $f_v$.
(2) Multi-Sequence Features $M_{multi} = \big \| _{v \in V} f_v(l_v) \big \| $ represents the concatenated features from all sequences for a particular lesion. This captures the combined information across all sequences. By incorporating both single-sequence and multi-sequence features, the model aims to capture the individual characteristics of each sequence and the relationships between them, potentially leading to improved breast lesion classification on MRI data.

\subsection{GIIM in the Missing-view Context}
To address the challenge of missing views, our approach first generates missing features and then uses them within multi-view methodologies. This exploration aims to identify suitable multi-view techniques and establish a meaningful representation of features for the missing views. Let $\mathcal{F}_v=\{\mathcal{F}_1, \mathcal{F}_2,..,\mathcal{F}_n\}$ denote the feature map extracted from the single-view model, where $\mathcal{F}_v^a \in \mathcal{F}_v$ denotes the feature maps representing available views. The goal is to generate missing view feature maps $\mathcal{F}_{v}^{m}$ from $\mathcal{F}_v^a$ with $\mathcal{F}_v^a \cup\mathcal{F}_v^m=\mathcal{F}_v$.

\subsubsection{Missing view:} \textbf{Constant} defines the feature vector for the missing view as a constant zero vector $F_v^{1 \times c} = [0.0]^{1 \times c}$ 

\subsubsection{Missing view:} \textbf{Learnable} treats the feature vector as a set of learnable parameters that are optimized during training. The final vector is produced by pooling a randomly initialized learnable tensor and then normalizing it using the Frobenius norm $\widehat{F}_v^{1 \times c} = \frac{\widehat{F}_v}{\|\widehat{F}_v\|_{F}}$

\subsubsection{Missing view: RAG-based}
We additionally applied a retrieval-augmented strategy focused on imputation rather than synthesis. If the feature was absent, we merged the available features, computed similarity across the dataset, identified the entry with the highest similarity, and used its corresponding missing features to populate the missing field. To calculate the missing feature $\mathcal{F}_a^m$, we can use the following formula:

\begin{equation}
    \begin{aligned}
    \mathbf{x}      &= 
    \begin{bmatrix}
        \mathcal{F}_v^{a,1} \\[2pt]
        ...\\
        \mathcal{F}_v^{a,k}
    \end{bmatrix}, \quad
    \mathbf{x}_i  = 
    \begin{bmatrix}
        \mathcal{F}_{v,db}^{a,1} \\[2pt]
        ...\\
        \mathcal{F}_{v,db}^{a,k}
    \end{bmatrix}, \quad
    s_i      &= 
    \frac{\mathbf{x}^{\mathsf T}\mathbf{x}_i}
         {\lVert \mathbf{x}\rVert_2 \,\lVert \mathbf{x}_i\rVert_2} \qquad \\[2pt]
    i^{\star}      &= \underset{i}{\arg\max}\; s_i \quad , \quad \mathcal{F}_v^m = \mathcal{F}^m_{v_{\,j^{\star}}}
    \end{aligned}
\end{equation}

Available features in $\mathcal{F}_v^a$ are stacked column vectors $\mathbf{x}$ (query) and $\mathbf{x}_i$ (database). Calculate the similarity $s_i$ for each pair $\bigl(\mathbf{x},\mathbf{x}_i\bigr)$, identify the index $i^{\star}$ with the highest similarity, and copy the corresponding missing value $\mathcal{F}^m_{v_{\,j^{\star}}}$ into $\mathcal{F}_v^m$.
\subsubsection{Missing view: Covariance-based} Besides, we propose a method to impute missing features by leveraging a covariance-based similarity metric. The method begins by constructing a feature space from a database of $k$ complete available views. This space is defined by a set of vectors $\mathcal{\Tilde{F}}_v^a = \{ \Delta_1, \Delta_2, \dots, \Delta_k \}$, where each vector $\Delta_i = \mathcal{F}_v^{a,1}-\sum_{j=2}^{k} \mathcal{F}_{v}^{a,j} \in \mathbb{R}^{c}$ quantifies the difference among available features for a given sample $i=1, \dots, n$. A covariance matrix $\mathbf{\Sigma} \in \mathbb{R}^{c \times c}$ is then computed over the feature space $\mathcal{F}$ to capture the statistical relationships between the feature components:
\begin{equation}
    \mathbf{\Sigma}=\frac{1}{n-1}\sum_{i=1}^{n}(\Delta_i-\mu)(\Delta_i-\mu)^T
\end{equation}
where $\mu \in \mathbb{R}^{c}$ is the mean vector of the feature space $\mathcal{F}$.
To generate the missing feature $\mathcal{F}_v^m$ for a query sample, its corresponding available feature difference vector, $\Delta^q$, is first computed. A covariance similarity score, $s_j$, is then calculated between this query vector and each vector $\Delta_j$ in the database. The index $j^{\star}$ of the database sample with the highest similarity is identified, and its delay feature is adopted, as follows:
\begin{equation}
    \begin{aligned}
        s_j &= (\Delta^q)^T \mathbf{\Sigma} \Delta_j \\
        j^{\star} &= \underset{j}{\arg\max}\; s_j, \quad \text{then } \mathcal{F}_v^m = \mathcal{F}^m_{v_{\,j^{\star}}}
    \end{aligned}
\end{equation}
This process imputes missing value by borrowing from the most statistically similar complete sample in feature space.
\section{Experiments}
\label{sec:experiment}

\begin{table*}[h]
\centering
\resizebox{\textwidth}{!}{
\begin{tabular}{ |l||l|l|l||l|l|l||l|l|l|| }
\hline
\multicolumn{1}{|c||}{} & \multicolumn{3}{c||}{\textbf{Liver}} & \multicolumn{3}{c||}{\textbf{VinDr-Mammo}} & \multicolumn{3}{c||}{\textbf{BreastDM}} \\
\hline
\multicolumn{1}{|c||}{\textbf{Type}} & \textbf{Methods} & Acc (\%) & AUC (\%) & \textbf{Methods} & Acc (\%) & AUC (\%) & \textbf{Methods} & Acc (\%) & AUC (\%) \\
\hline
\hline
\multirow{3}{*}{\rotatebox[origin=c]{90}{\textbf{Single}}} & Arterial & 65.91 & 82.78 & CC & 62.56 & 77.81 & Pre-contrast & 82.97 & 82.94 \\ \cline{2-10}
& Venous & 68.64 & 84.91 & MLO & 61.57 & 78.01 & Post-contrast & 72.34 & 75.29 \\ \cline{2-10}
& Delay & 66.82 & 82.83 & & & & Subtraction & 76.59 & 77.84 \\
\hline
\hline
\multirow{4}{*}{\rotatebox[origin=c]{90}{\textbf{Multi}}} & NN-based & 75.45 & 89.09 & NN-based & 67.48 & 82.21 & NN-based & 80.85 & 87.35 \\ \cline{2-10}
& ML-based & 73.63 & 88.00 & ML-based & 66.87 & 80.86 & ML-based & 82.98 & 79.41 \\ \cline{2-10}
& Attention-based & 73.41 & 88.53 & Attention-based & 68.09 & 81.00 & Attention-based & 85.1 & 76.37 \\ \cline{2-10}
& \textbf{GIIM (ours)} & \textbf{78.20} & \textbf{91.05} & \textbf{GIIM (ours)} & \textbf{71.17} & \textbf{82.54} & \textbf{GIIM (ours)} & \textbf{87.23} & \textbf{89.02} \\
\hline
\end{tabular}
}
\caption{Performance metrics in terms of accuracy and AUC of different single- and multi-phase methods evaluated on the liver tumor test set, the VinDr-Mammo dataset and the BreastDM dataset.}
\label{tab:liverresuls}
\end{table*}

This section first presents the datasets used for model evaluations and then describes experimental results and analysis.
\subsection{Datasets}
\subsubsection{Liver Tumor Dataset}

Our method's performance is evaluated on a private dataset comprising 920 abdominal CT examinations conducted between 2016 and 2022. These exams originate from 877 patients, ranging in age from 5 to 96 years, and contain four distinct 3D CT images corresponding to different phases of contrast enhancement: non-contrast, arterial, venous, and delayed phases. We stratify the dataset into two groups using label stratification: a training set (659 cases) and a test set (261 cases). Our classification categorizes lesions into four distinct classes based on their pathological characteristics: benign, ambiguous, malignant other than HCC, and HCC tumors. Benign lesions encompass a variety of liver tumors such as focal nodular hyperplasia (FNH), cyst, hemangioma, abscess, differential diagnosis (DD), and adenoma. The ambiguous label contains two types of liver tumors, LIRADS 3 and ambiguous DD. The malignant group contains three types of tumors: intrahepatic cholangiocarcinoma (ICC), metastases, and malignant DD. Detailed dataset statistics are listed in Table ~\ref{tab:liverdataset}.

\subsubsection{VinDr-Mammo Dataset}

The VinDr-Mammo dataset \cite{vindrmammo} is a comprehensive benchmark for full-field digital mammography, containing 5,000 four-view examinations with 20,000 scans. Each scan is annotated with breast-level BI-RADS and findings information, making it the most extensive public resource for BI-RADS assessment. The mammography images were acquired from H108 and HMUH sources. In our approach, we extract ROIs representing lesions from mammograms for the BI-RADS classification task. Each lesion is annotated with an internal BI-RADS label (3-5). A sample consists of CC and MLO views, each containing multiple lesion ROIs. The global label for the sample is the highest pathology, denoted as $y_G = \max([y_{l=1}^{L_{CC}} || y_{l=1}^{L_{MLO}}])$. Model performance is evaluated at the sample level. 
The dataset comprises 783 samples (1,930 lesions), split into a \textbf{training set} of 620 samples (1,532 lesions) and a \textbf{test set} of 163 samples (398 lesions). The label distribution for BI-RADS \{3, 4, 5\} is $N_{train} = \{264, 271, 85\}$ and $N_{test} = \{71, 71, 21\}$ for the training and test sets, respectively.

\begin{table*}[h]
\centering
\resizebox{\textwidth}{!}{
\begin{tabular}{|c|l||c|c|c|c|c||c|c|c|c|c|}
\hline
\multicolumn{2}{|c||}{} & \multicolumn{5}{c||}{\textbf{Test on 100\% missing-view}} & \multicolumn{5}{c|}{\textbf{Test on full view}} \\
\cline{3-12}
\multicolumn{2}{|c||}{\multirow{-2}{*}{\textbf{Method}}} & $\eta=0.0$ & $\eta=0.2$ & $\eta=0.5$ & $\eta=0.7$ & $\eta=1.0$ & $\eta=0.0$ & $\eta=0.2$ & $\eta=0.5$ & $\eta=0.7$ & $\eta=1.0$ \\
\hline
\hline
\multirow{7}{*}{\rotatebox[origin=c]{90}{\textbf{Liver Dataset}}} & NN-based & 70.00 & 72.50 & 70.23 & 71.82 & 72.50 & 75.45 & 73.41 & 71.36 & 71.59 & 72.50 \\
& ML-based & 71.59 & 72.95 & 72.05 & 72.27 & 72.73 & 73.63 & 73.86 & 72.73 & 73.18 & 72.73 \\
& Attention-based & 67.50 & 73.18 & 71.36 & 67.95 & 72.73 & 73.41 & 71.82 & 72.95 & 64.09 & 67.50 \\
& GIIM (constant) & 72.27 & 73.86 & 72.73 & \textbf{74.09} & \textbf{73.41} & \textbf{78.20} & 75.68 & 73.41 & 72.50 & 73.86 \\
& GIIM (learnable) & 73.64 & 73.86 & \textbf{73.86} & 72.95 & 70.91 & \textbf{78.20} & 74.09 & \textbf{76.82} & 76.82 & 65.45 \\
& GIIM (RAG-based) & \textbf{74.31} & 72.73 & 72.95 & 71.81 & 72.50 & \textbf{78.20} & 74.55 & 75.91 & 76.14 & \textbf{76.82} \\
& GIIM (Covariance) & 70.91 & \textbf{74.32} & 72.95 & 73.18 & 72.73 & \textbf{78.20} & \textbf{78.41} & 75.91 & \textbf{77.95} & 74.31 \\
\hline
\hline
\multirow{7}{*}{\rotatebox[origin=c]{90}{\textbf{VinDr-Mammo}}} & NN-based & 63.19 & 63.19 & 63.19 & 64.42 & 63.80 & 67.48 & 63.19 & 63.80 & 65.03 & 62.58 \\
& ML-based & 58.28 & 61.96 & 63.80 & 63.19 & 62.58 & 66.87 & 65.03 & 63.80 & 63.80 & 62.58 \\
& Attention-based & 63.19 & 65.03 & 65.64 & 65.64 & 66.26 & 68.09 & 63.19 & 63.80 & 65.03 & 63.04 \\
& GIIM (constant) & 61.35 & 65.03 & \textbf{66.26} & 66.26 & \textbf{66.26} & \textbf{71.17} & 66.87 & 65.03 & 65.03 & \textbf{66.26} \\
& GIIM (learnable) & \textbf{63.19} & \textbf{65.03} & 64.42 & \textbf{66.26} & 65.64 & \textbf{71.17} & 64.42 & 63.19 & 61.35 & 60.74 \\
& GIIM (RAG-based) & 59.51 & 60.74 & 62.58 & 61.96 & 61.96 & \textbf{71.17} & 65.64 & \textbf{66.87} & \textbf{68.71} & \textbf{66.26} \\
& GIIM (Covariance) & 61.35 & 62.58 & 59.51 & 63.80 & 63.80 & \textbf{71.17} & \textbf{67.48} & 65.64 & 66.26 & \textbf{66.26} \\
\hline
\end{tabular}
}
\caption{Comparison in terms of classification accuracy with missing view rate $\eta = [0.0, 0.2, 0.5, 0.7, 1.0]$ on Liver and VinDr-Mammo datasets.}
\label{tab:missingresult}
\end{table*}

\subsubsection{BreastDM Dataset}
BreastDM \cite{breastdm} is a dynamic contrast-enhanced magnetic resonance imaging (DCE-MRI) dataset curated for advancing research in breast lesion classification, particularly focusing on malignant and benign differentiations. The dataset encompasses a total of 232 cases, comprising 147 malignant and 85 benign lesions. Each case within the dataset includes three distinct types of sequences: pre-contrast, post-contrast, and subtraction.

\subsection{Experiment Results and Analysis}
\subsubsection{Multi-view Experiments}

In modern medical image analysis, multi-view classification techniques can be broadly categorized by their core methodology. These include traditional ML methods that excel with structured data, CNN based approaches that focus on local spatial patterns, Attention/Transformer-based models designed to capture global dependencies, and GNN based techniques that explicitly model relationships between different entities. To comprehensively evaluate GIIM, we compare its performance against established techniques from these other domains. To ensure a \textbf{fair comparison}, all methods share the same core feature extraction backbone, which was originally designed for single-view models. This isolates the comparison to the effectiveness of the multi-view learning technique itself. The classifiers we evaluate against are:
\begin{itemize}
    \item \textbf{NN-based}: A classifier that utilizes two fully connected layers to process the extracted features.
    \item \textbf{ML-based}: The LightGBM classifier, a powerful gradient-boosting framework for structured feature data, as implemented by ~\cite{nguyen2022novel}.
    \item \textbf{Attention-based}: Phase attention modules introduced by ~\cite{wang2022phase}, which use an attention mechanism to weigh the importance of features from different views.
\end{itemize}

The performance of these approaches on the private liver dataset, the VinDr-Mammo dataset and BreastDM dataset is presented in Table~\ref{tab:liverresuls}. It can be observed from Table \ref{tab:liverresuls}  that multi-view approaches achieved significant higher performances, showcasing large margin of consistent improvements of approximately 12\% in accuracy and 8.3\% in AUC on the in-house dataset, 7.8\% in accuracy and 4.8\% in AUC on VinDr-Mammo, as well as 5\% in accuracy and 7\% in AUC on BreastDM when compared to single-view methods. This result can justify the fact that multi-view medical imaging provides a complete perspective of lesions, capturing insights from mammogram observation angles to different observation timeframes in CT scans and breast MRI.

Among the multi-view approaches, GIIM model consistently outperforms other multi-view methods, with an 3\% improvement in accuracy and 2\% increment in AUC on the liver tumor dataset. For the mammogram dataset, our GIIM achieves the highest accuracy among the tested methods. The same observation can be drawn from results obtained on the BreastDM dataset, in which the proposed GIIM obtained highest accuracy of 87.23\% and AUC of 89.02\%.

\subsubsection{Missing-view Experiments}

In medical analysis, missing views present a significant challenge for accurate diagnosis. This issue occurs for various reasons, including routine clinical protocols that may not require all views (e.g., using only arterial and venous phases for FLL classification) , as well as technical errors or patient refusal. Such incomplete data can severely degrade or even deactivate the predictive performance of multi-view models. To assess the influence of incomplete view data on model performance, we trained models on datasets with varying rates of missing views ($\eta$) and then evaluated on two distinct test sets: \textbf{Full-view Test}, where all samples have complete views, and \textbf{Miss-view Test}, where all samples are missing one view. For FLLs classification, we consider the scenario where the Delay phase is missing, whereas for BI-RADS classification, we assume the CC view is missing.

First, we compared our basic GIIM model (using a constant vector for missing views) against other multi-view methods. As shown in Table~\ref{tab:missingresult}, \textbf{GIIM (constant)} model consistently and significantly outperforms the NN-based, ML-based and Attention-based methods in both the full view and miss view test sets across two datasets. This demonstrates the fundamental advantage of explicitly modeling inter- and intra-lesion relationships for multi-view diagnosis.

We then investigated four different methods for handling missing data within GIIM framework: constant, learnable, RAG-based, and Covariance-based. A clear overall trend is that all methods perform better on the \textbf{Full-view Test} than the \textbf{Miss-view Test} around 4\% in Liver dataset and 5\% in VinDr-Mammo. This confirms the intuitive idea that having complete data at the time of diagnosis leads to better accuracy. Interestingly, there is a trade-off between performance on the two test sets. On \textbf{Full-view Test}, RAG-based and Covariance methods often achieve the best results. This suggests that these methods learn to generate a feature vector with a similar distribution to a real view's features, which is beneficial when a full set of views is available. On \textbf{Miss-view Test}, however, the simple constant vector method is often more effective. A likely explanation is that using a unique, constant vector makes the missing node obvious to the graph, causing the model to learn to rely more heavily on the available nodes.

\section{Conclusion}
\label{sec:conclusion}
 This paper introduces GIIM, a novel MHGs-based approach for medical image classification that leverages both intra- and inter-view information inherent in several imaging modalities. Through extensive experimental validation on both internal and public datasets, our proposed method yields significantly enhanced diagnostic precision compared to several existing single-phase and other multi-view techniques. Furthermore, for the cases of incomplete viewing data, our GIIM demonstrates remarkable performance and robustness compared to existing multi-view approaches. Our study opens up a promising approach to constructing and developing an efficient CADx, that can work well even when lacking input information and thus potentially benefit practical clinics. Our ongoing work involves exploring novel MHGs architectures for medical imaging analysis, specifically focusing on exploring lesion relationships and integrating expert knowledge to construct powerful MHGs.


\end{document}